\PassOptionsToPackage{dvipsnames}{xcolor}
\PassOptionsToPackage{dvipsnames}{color}
\documentclass[runningheads]{llncs}
\usepackage[T1]{fontenc}
\usepackage{graphicx,verbatim}
\usepackage{amsmath}
\usepackage[ruled,vlined]{algorithm2e}
\PassOptionsToPackage{usenames,dvipsnames}{xcolor}
\usepackage{marginnote}

\usepackage[usenames,dvipsnames]{color}
\DefineNamedColor{named}{Orange}{cmyk}{0.01,0.5,0.97,0}
\DefineNamedColor{named}{Red}{cmyk}{0,0.999,0.99,0}
\DefineNamedColor{named}{RedOrange}{cmyk}{0,0.77,0.87,0}
\DefineNamedColor{named}{LimeGreen}{cmyk}{0.50,0,1,0}
\DefineNamedColor{named}{Green}{cmyk}{1,0,1,0}
\DefineNamedColor{named}{ForestGreen} {cmyk}{0.91,0,0.88,0.12}
\DefineNamedColor{named}{Emerald}{cmyk}{1,0,0,0}
\DefineNamedColor{named}{DarkEmerald}{cmyk}{1,0,0,0.4}
\newif\ifshowcomments
\showcommentsfalse 
\newif\ifshowinlinecomments
\showinlinecommentstrue 
\ifshowinlinecomments

\def\Gc#1{\textcolor{Emerald}{\small \textsf{(G: #1)}}}
\def\Pc#1{\textcolor{red}{\small \textsf{(P: #1)}}}
\def\Gcm#1{\textcolor{Emerald}{$^\star$}\marginnote{\textcolor{Emerald}{\scriptsize \textsf{G:#1}}}}
\else

\def\Gc#1{}
\def\Pc#1{}
\def\Gcm#1{}
\fi

\begin{document}

\title{AREPAS: Anomaly Detection in Fine-Grained Anatomy with Reconstruction-Based Semantic Patch-Scoring}

\titlerunning{AREPAS}

 \author{Branko Mitic\inst{1,2,3}\orcidID{0009-0006-7838-2370} \and
 Philipp Seeb\"ock\inst{1,2}\orcidID{0000-0001-5512-5810} \and
 Helmut Prosch\inst{3}\orcidID{0000-0002-6119-6364} \and
 Georg Langs\inst{1,2,3}\orcidID{0000-0002-5536-6873}}

\authorrunning{B. Mitic et al.}

\institute{$^1$ Computational Imaging Research Lab,  Department of Biomedical Imaging and Image-guided Therapy, Medical University of Vienna\\$^2$
Comprehensive Center for Artificial Intelligence in Medicine, \\ Medical University of Vienna \\$^3$
Christian Doppler Laboratory for Machine Learning Driven Precision Imaging, Department of Biomedical Imaging and Image-guided Therapy, \\ Medical University of Vienna, Austria \\
\email{branko.mitic@meduniwien.ac.at, georg.langs@meduniwien.ac.at} \\
\url{https://www.cir.meduniwien.ac.at}
\\}
    
\maketitle              

\begin{abstract}

Early detection of newly emerging diseases, lesion severity assessment, differentiation of medical conditions and automated screening are examples for the wide applicability and importance of anomaly detection (AD) and unsupervised segmentation in medicine. Normal fine-grained tissue variability such as present in pulmonary anatomy is a major challenge for existing generative AD methods. Here, we propose a novel generative AD approach addressing this issue. It consists of an image-to-image translation for anomaly-free reconstruction and a subsequent patch similarity scoring between observed and generated image-pairs for precise anomaly localization. We validate the new method on chest computed tomography (CT) scans for the detection and segmentation of infectious disease lesions. To assess generalizability, we evaluate the method on an ischemic stroke lesion segmentation task in T1-weighted brain MRI. Results show improved pixel-level anomaly segmentation in both chest CTs and brain MRIs, with relative DICE score improvements of +1.9\% and +4.4\%, respectively, compared to other state-of-the-art reconstruction-based methods.


\end{abstract}


\section{Introduction}

\begin{figure}[t]
  \centering
  \includegraphics[width=1\linewidth]{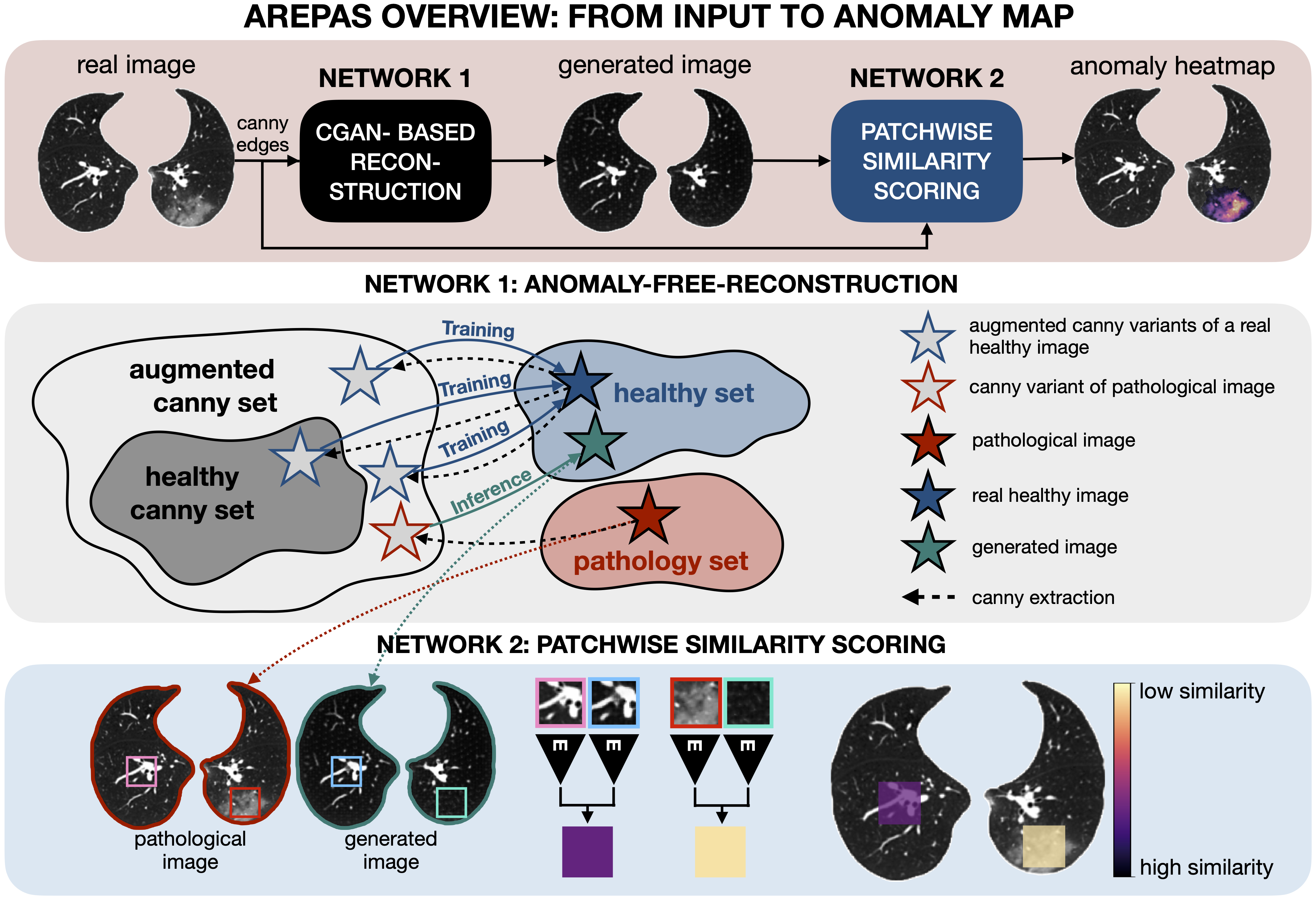}
  \caption{AREPAS uses two networks: (1) The reconstruction network is designed to reconstruct the real image without anomalies from the canny-edge representation of the observed image. (2) The second network samples a set of random patch-pairs from same locations (label 1) and differing locations (label 0) in the reconstructed and real images. These patch-pairs, with corresponding labels, are used to train a Siamese network with a contrastive loss. During inference we compute an ensemble of patch-level similarity scores from identical locations in the real and reconstructed image-pairs and aggregate them to obtain an anomaly heat-map.
}  
\label{fig: main}
\end{figure}

Clinical imaging is essential for the rapid identification of rare or emerging diseases and the effective treatment of patients \cite{fang2021covid}. Most notably, it has played a crucial role in detecting and analyzing outbreaks of respiratory diseases that later became pandemics, including SARS, MERS, H1N1, and COVID-19 \cite{Wilder-Smith2020-rz}. In this area, traditional diagnostic methods relying on supervised learning face significant limitations, as large, annotated datasets are unavailable for scarce or previously unseen diseases. To address this, anomaly detection has emerged as a powerful alternative. It is capable of detecting and segmenting anomalies without the need for labeled training data. Instead, training is exclusively performed on non-anomalous samples.

\paragraph{Related work} In medical anomaly detection, a range of methods have been proposed, including approaches involving generative networks in retinal imaging \cite{schlegl2017unsupervised,schlegl2019f,seebock2018unsupervised} or generative diffusion-based networks in brain MRIs \cite{liang2024itermask,bercea2024diffusion,bercea2023mask,wyatt2022anoddpm}. Conversely, there has been a limited number of publications focusing on anomaly detection and localization in CT-scans and chest X-rays related to lung conditions \cite{yadav2021lung,nakao2021unsupervised,kim20223d}. One reason for this imbalance is that generative models struggle to accurately reconstruct healthy areas in medical images, including fine-grained and intricate structures such as pulmonary anatomy. To address this problem it may be necessary to find a balance during the mapping of anomalous to non-anomalous tissue appearance without sacrificing accurate reconstruction of healthy tissue. Recently, initial progress targeting this problem has been made~ \cite{liang2024itermask,bercea2024diffusion}, but limitations remain as models continue to struggle to generalize across domains or handle diverse lesion types effectively.

\paragraph{Contribution} To address these limitations we introduce a new method for pixel-level anomaly detection. The approach uses edges  to condition the reconstruction on structural detail, and replaces a mere residual with a scoring network that can distinguish between normally occurring deviations, and those that indicate actual anomalies. First, a conditional generative model translates augmented canny-edge representations of an input image to an estimate of its non-anomalous equivalent. In a second step, instead of relying on pixel-level comparison of the real input image and the generated non-anomalous version to detect anomalies, we introduce a scoring function for patch-wise similarity assessment and semantic anomaly localization. We validate our method on two publicly available datasets; MosmedData chest CTs \cite{morozov2020mosmeddata} and ATLAS v2.0 brain T1-weighted MRIs \cite{liew2022large}, substantiating the applicability of the proposed method across domains.


\section{Method}

We aim to detect the presence and location of anomalies in an image $\boldsymbol{I}^{real}_i$, using only non-anomalous data during model training. Our approach consists of two steps (Fig.~\ref{fig: main}). 
In \textbf{step 1} we generate a non-anomalous reconstruction $\boldsymbol{I}^{rec}_i$ of the real input $\boldsymbol{I}^{real}_i$ using edge detection as intermediate processing of the input image. 
In \textbf{step 2} we compute an ensemble of similarity scores between patch-pairs in the reconstructed $\boldsymbol{I}^{rec}_i$ and real image $\boldsymbol{I}^{real}_i$ and aggregate them in a heat-map $\boldsymbol{A}_i$ to localize anomalies in the image.

To this end, we train two networks. The first network \( N^{rec} \) generates the reconstruction from an edge-image (step 1). The second network, \( N^{siam} \) performs patch-level similarity scoring between the real image and its reconstruction (step 2). The final segmentation $\boldsymbol{A}^{final}_i$ is computed by multiplying $\boldsymbol{A}_i$  element-wise with the pixel residual between real $\boldsymbol{I}^{real}_i$ and reconstructed image $\boldsymbol{I}^{rec}_i$,
\begin{equation}
    \boldsymbol{A}^{final}_i=|\boldsymbol{I}^{real}_i-\boldsymbol{I}^{rec}_i| \circ \boldsymbol{A}_i
\end{equation}

\subsection{Anomaly-free reconstruction: $\boldsymbol{I}^{real}\mapsto \boldsymbol{I}^{rec}$}\label{sec: 2.1}

\paragraph{Training} We train a conditional generative adversarial network inspired by pix2pixHD~\cite{wang2018high} $N^{rec}: \boldsymbol{\mathcal{I}}^{canny}\mapsto \boldsymbol{I}^{rec}$ on a set of non-anomalous image samples. $N^{rec}$ reconstructs anomaly-free images from an edge-image obtained with canny edge extraction~\cite{canny1986computational} $C$ from the input image, yielding the overall model $G(\boldsymbol{I}^{real}) = N^{rec}(C(\boldsymbol{I}^{real}))$ with $G: \boldsymbol{I}^{real}\mapsto \boldsymbol{I}^{rec}$. $N^{rec}$ is trained on pairs of augmented edge-images and corresponding real images $\langle \boldsymbol{\mathcal{I}}^{canny}_{i,j},\boldsymbol{I}^{real}_i\rangle$, where $j$ is the index of augmented versions of the true edge-image. We augment $\boldsymbol{\mathcal{I}}^{canny}_{i}$ by randomly copy-pasting image regions.  

Details of the model and augmentation procedure are described in section \ref{sec: implementation details}.

\paragraph{Inference} During inference on a new image $\boldsymbol{I}^{real}_{new}$, $G$ yields the corresponding reconstruction $\boldsymbol{I}^{rec}_{new}$, an estimate of an anomaly-free reconstruction of the input image.

\subsection{Semantic anomaly localization: $\langle\boldsymbol{I}^{real}, \boldsymbol{I}^{{rec}}\rangle\mapsto \boldsymbol{A}$}\label{sec: 2.2}

\paragraph{Training} To generate an anomaly heat-map $\boldsymbol{A}$ we train a convolutional Siamese network with a contrastive loss~\cite{bromley1993signature,chopra2005learning,koch2015siamese} on normal training data to compute patch-level similarity scores between observed and reconstructed patches extracted at corresponding positions $\langle \boldsymbol{p}^{real}_k,\boldsymbol{p}^{{rec}}_k\rangle \mapsto a_k$. These patches constitute positive image-pairs (label $y_k=1$). Negative pairs consist of patches extracted at  different locations in the reconstructed and real images (label $y_k=0$). Each patch-pair is independently processed by two identical encoder networks of the siamese network with shared weights. The Euclidean distance $d_k$ between the resulting feature vectors $v^{real}_k$ and $v^{rec}_k$ is calculated and scaled from zero to one with a sigmoid activation function resulting in the similarity score $a_k$. We train the network with a contrastive loss   
$$
L = \frac{1}{b} \sum_{k=1}^{b} \left( (1 - y_k) \cdot a_k^2 + y_k \cdot\max(0, 1 - a_k)^2 \right),
$$
where $y_k$ is the ground truth label, $d_k$ the Euclidean distance between feature vectors $v^{\mathrm{real}}_k$ and $v^{\mathrm{rec}}_k$, and $b$ the batch size.

\paragraph{Inference} During inference on a pair of observed and reconstructed images $\langle\boldsymbol{I}^{real}_i, \boldsymbol{I}^{{rec}}_i\rangle$ we sample patch-pairs at positions across the image, evaluate their similarity score $a_k$, and reassamble them at their positions to the full resolution heat-map $\boldsymbol{A}_i$ scaled between zero and one. 

\subsection{Rationale of the approach}

In the first step, we generate a reconstruction of the observed real image from its canny-edge representation \cite{canny1986computational} using a generative model. Edge-maps preserve structural information while removing pixel intensities, forcing the model to rely on learned anatomical priors, and emphasizing fine detail. To avoid overfitting to a rigid translation of the edge-representation we apply augmentations during training~\cite{ghiasi2021simple}. The network is trained on several corrupted edge-maps paired with the same target image, encouraging invariance to edge irregularities. The intuition is to artificially enhance the input distribution encompassing the full scope of potentially anomalous canny-edge representations, and encouraging the generator to yield images consistent with the real data distribution rather than learning a rigid input-to-output mapping. The anomaly-free reconstruction section in Fig.\,\ref{fig: main} illustrates this idea. Since reconstructions may slightly deviate from their corresponding true images even if no anomalies are present, we introduce a network that learns which differences of the reconstructed and corresponding real image are normal and which indicate anomalies. This step is essential for two reasons. First, highly variable fine structures in images are difficult to reconstruct and may result in higher pixel-level errors in areas where no anomalies are present. Second, as demonstrated in \cite{mousakhan2023anomaly}, a pixel-level residual map may not carry enough information to distinguish between an anomalous and non-anomalous region when pixel intensities of these regions are similar. 


\section{Experimental Setup}

\paragraph{Data and preprocessing} Experiments are performed on MosmedData chest CTs \cite{morozov2020mosmeddata} and ATLAS v2.0 ischemic brain lesion T1-weighted MRIs \cite{liew2022large}. The CT data includes healthy patients and patients mildly affected by COVID-19. The networks are trained on exclusively healthy slices from patient-level splits. Further information regarding data is contained in Tab. ~\ref{tab:data_table}. We clip the CTs from $-1000$ HU to $0$ HU and scale the values to the range between $-1$ and $1$. Then we segment the lungs using a publicly available method~\cite{hofmanninger2020automatic} and set the background to zero. We select adjacent slices covering $90\%$ of the lung volume in healthy scans to avoid training on slices with minimal lung tissue. For infected cases in test and validation sets, we include one slice with the largest manually segmented infection area for each patient. Note that in some cases, the largest region may still be small or faint, even though it's the most prominent lesion present in that patient. All images are square cropped to a bounding box of the lung to remove as much background as possible and resized to $256 \times 256$ pixels. 

In a second experiment, we perform training and evaluation on mid-axial slices of brain imaging data, with 98th percentile clipping and square padding. We train on 582 T1-weighted slices from IXI~\cite{IXI} and 196 mid-axial slices from ATLAS \cite{liew2022large} without labeled anomalies ensuring no patients overlap between training and test/validation sets. For testing we use 438 scans with expertly annotated lesion masks. Slices with large, unannotated artifacts are excluded. This is consistent with the experimental setup used in~\cite{bercea2024diffusion}.

\paragraph{Evaluation metrics} We evaluate the performance of our model with pixel-level area under precision recall curve (AUPRC). For an anomaly map threshold determined using the validation set we report DICE similarity coefficient, precision and recall. We compare resulting metrics to several state-of-the-art approaches: IterMask \cite{liang2024itermask}, THOR (simplex noise) \cite{bercea2024diffusion}, AnoDDPM (simplex noise) \cite{wyatt2022anoddpm}, AutoDDPM (gaussian noise) \cite{bercea2023mask}.

\paragraph{Ablation study and patch size comparison} We perform an ablation study to demonstrate the effectiveness of the novel network components: (1) copy-paste augmentations to edge-inputs, and (2) semantic patch-based anomaly localization. We compare the DICE score across a range of patch sizes for similarity scoring as a hyperparameter.

\paragraph{Implementation details}\label{sec: implementation details} The generator follows an encoder--residual--decoder structure with two convolutional downsampling layers, nine $3{\times}3$ ResNet blocks (with dropout $p{=}0.5$), and two transposed convolutions for upsampling. Input/output use $7{\times}7$ convolutions with reflection padding. Instance normalization~\cite{ulyanov2016instance} and ReLU activations are used throughout. The discriminator is a patch-based network with five spectral-normalized~\cite{miyato2018spectral} convolutional layers, LeakyReLU ($\alpha{=}0.2$), and strided downsampling on concatenated input--target pairs. Adversarial training uses binary cross-entropy loss with label smoothing (real $= 0.9$). Generator loss combines adversarial, L1 ($\lambda_{\mathrm{L1}}{=}100$), and perceptual loss ($\lambda_{\mathrm{VGG}}{=}1$) using \texttt{block3\_conv3} features from VGG-19~\cite{simonyan2014very}. Discriminator loss includes a gradient penalty \cite{gulrajani2017improved} ($\lambda_{\mathrm{GP}}{=}1$). Optimization uses Adam~\cite{kingma2014adam} with learning rate $2{\times}10^{-4}$, $\beta_1{=}0.5$, batch size 1, for 10 epochs.
Edge-input augmentations involve randomly swapping smooth, irregular patches (1--33\% of image area), shaped via Gaussian-blurred noise and contour smoothing. Up to 20 augmentations per image are applied, each with up to 10 copy-paste operations. Canny thresholds are both set at 66\% of the Otsu threshold~\cite{otsu1975threshold}.

The Siamese encoders apply batch normalization, two convolutional layers (32 and 64 filters, $4{\times}4$ kernels, Tanh, same padding) with $2{\times}2$ average pooling, followed by flattening, batch normalization, and a fully connected layer (10 units, Tanh). The network is trained for 50 epochs with batch size 1024, selecting the best validation accuracy. 

The code is available at \url{https://github.com/cirmuw/arepas}.

\begin{table}[t!]
\begin{center}
\caption{Data overview. Experiments are evaluated on MosmedData chest CTs \cite{morozov2020mosmeddata} and ATLAS v2.0 ischemic brain lesion T1w-MRIs \cite{liew2022large}.}
\begin{tabular}{ r|c|c } 

 \textbf{Lung CT } & train healthy  & unhealthy   \\ 
  \hline
  \hline
slices & 5791 & 25 test|25 val.  \\ 
patients & 204 & 25 test|25 val.  \\ 
  \hline
  \hline
\textbf{Brain MRI } & train healthy & unhealthy \\
  \hline
  \hline
slices & 778 (196 ATLAS|582 IXI) & 219 test|219 val. \\ 
patients & 778 (196 ATLAS|582 IXI) & 219 test|219 val.  \\ 

\end{tabular}
\label{tab:data_table}
\end{center}
\end{table}


\section{Results}

\paragraph{Quantitative Results} 

Table~\ref{tab:results_table} summarizes the results for anomaly segmentation on the public MosmedData chest CT and ATLAS v2.0 brain T1-weighted MRI datasets. Our model achieves a mean DICE score of 0.638 for lung lesions and 0.31 for ischemic stroke lesions. This represents a relative performance improvement of 1.9\% for lungs and 4.4\% for brains compared to the best-performing state-of-the-art method in our experiments. Omitting semantic patch-scoring leads to a substantial relative performance drop (DICE) of 30.4\% (to 0.444) in lungs and 61.9\% (to 0.118) in brains. Additionally omitting edge-input augmentations further reduces DICE in lung data to 0.351, while it does not have a further effect in brain data (stays at 0.118).

\paragraph{Qualitative Results}

\begin{figure}[t]
  \centering
  \includegraphics[width=1\linewidth]{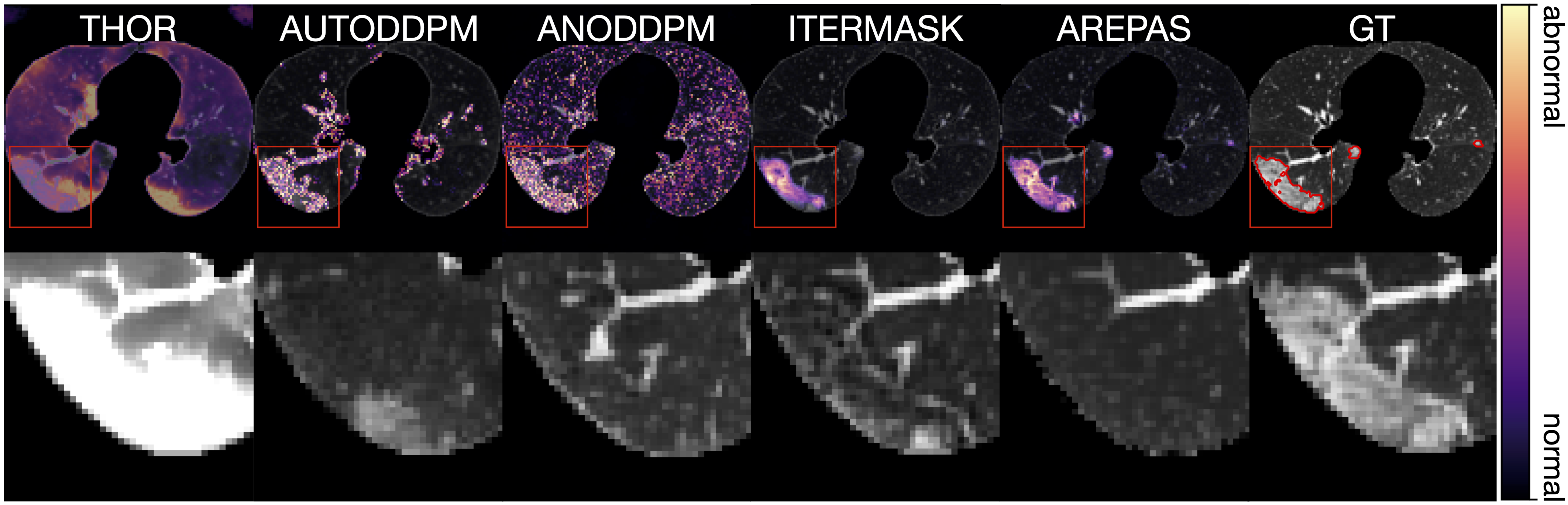}
  \caption{Qualitative anomaly segmentation (top row) and  zoomed in reconstruction results (bottom row) in Lung CTs. AutoDDPM is able to localize the lesion but fails to reconstruct the fine-grained lung vessels. AnoDDPM produces noisy anomaly maps. IterMask slightly under-segments and misses two small lesions. THOR appears not to translate to Lung CTs.
}  
\label{fig: qualitative results}
\end{figure}

\begin{figure}[t]
  \centering
  \includegraphics[width=1\linewidth]{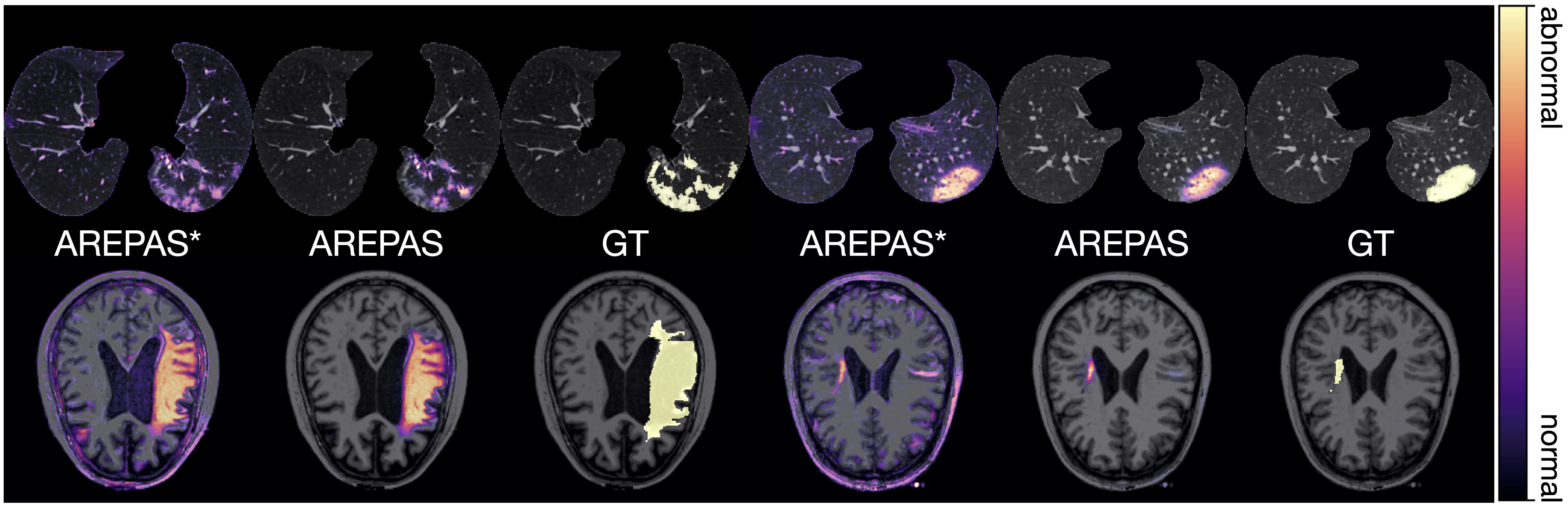}
  \caption{Qualitative ablation study results. The top row shows anomaly maps and ground truth masks for lung CTs with two lesions; the bottom row shows T1-weighted brain MRIs with large and small lesions. AREPAS* (without semantic patch-scoring) shows reduced segmentation quality, highlighting the importance of semantic patch-scoring.
}  
\label{fig: qualitative ablation}
\end{figure}

Fig.~\ref{fig: qualitative results} shows a representative qualitative result on lung CT data. Consistent with the quantitative results, our proposed method outperforms other state-of-the-art methods in anomaly localization and segmentation. Ablation results in Fig.~\ref{fig: qualitative ablation} illustrate the scope of improvement stemming from semantic patch-scoring. Competing methods exhibit noisy residual maps, poor reconstruction quality, and under-segmentation. By learning which deviations between the reconstructed and real images are normal, our method distinguishes true anomalies from reconstruction artifacts, thereby successfully addressing these limitations.

\begin{figure}[t]
  \centering
  \includegraphics[width=1\linewidth]{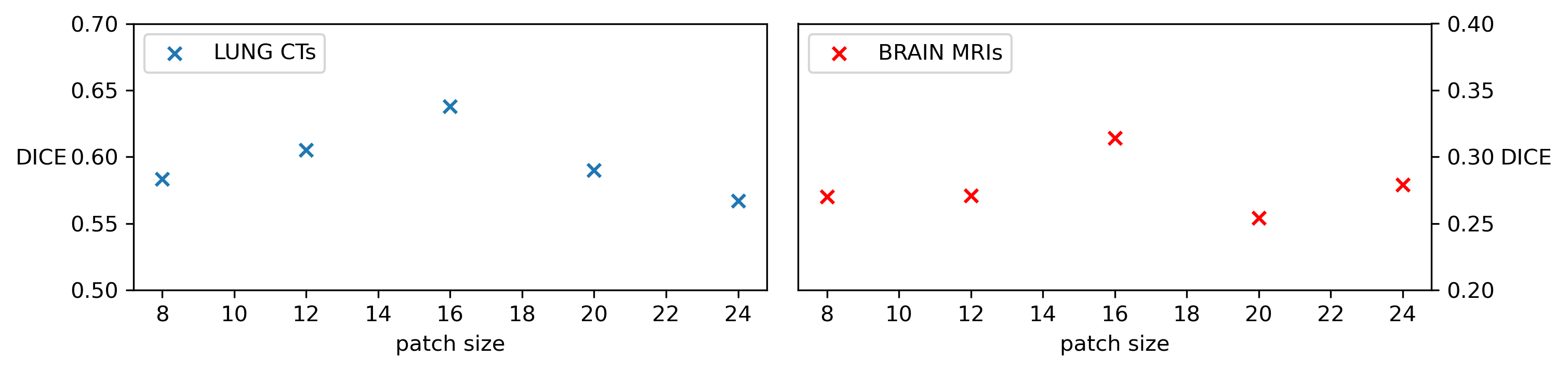}
  \caption{Hyperparameter study: semantic scoring patch size appears stable across various sizes with a subtle peak at $16 \times 16$ pixels for both modalities.
}  
\label{fig: hyperparameter}
\end{figure}

\paragraph{Influence of varying patch size for similarity scoring}
Fig.~\ref{fig: hyperparameter} demonstrates stable performance for patch sizes for similarity scoring ranging from $8 \times 8$ to $24\times 24$ pixels with a subtle peak at $16\times 16$.

\begin{table}[t!]
\centering
\caption{Quantitative results on lung CTs and brain MRIs. AREPAS* refers to our model without semantic patch-scoring. In AREPAS** we additionally omit edge-domain augmentations for ablation. Blue numbers indicate the relative performance improvement compared to the best state-of-the-art method. Red numbers indicate the relative performance decrease compared to our model. Approaches are sorted by DICE. }
\begin{minipage}[t]{0.62\textwidth}
\centering
{\fontsize{8pt}{10pt}\selectfont
\begin{tabular}{ r|c|c|c|c } 
\textbf{LUNG CTs} & DICE &  PR &  RE & AUPRC \\ 
  \hline
  \hline
THOR & 0.087$\pm 0.010$ {\color{red}(↓86.4\%)} & 0.045 & \textbf{0.902} & 0.038 \\ 
AutoDDPM & 0.174$\pm 0.015$ {\color{red}(↓72.7\%)} & 0.132 & 0.312 & 0.073 \\ 
AnoDDPM & 0.229$\pm 0.014$ {\color{red}(↓64.1\%)} & 0.187 & 0.362 & 0.154 \\ 
AREPAS** & 0.351$\pm 0.025$ {\color{red}(↓45.0\%)} & 0.329 & 0.485 & 0.348 \\ 
AREPAS* & 0.444$\pm 0.031$ {\color{red}(↓30.4\%)} & 0.417 & 0.601 & 0.487 \\ 
IterMask2 & 0.626$\pm 0.028$ {\color{red}(↓1.9\%)} & 0.564 & 0.725 & \textbf{0.700} \\ 
\hline
\textbf{AREPAS} & \textbf{0.638}$\pm 0.029$ {\color{blue}(↑1.9\%)} & \textbf{0.622} & 0.677 & 0.696 \\ 
\end{tabular}

}
\label{tab:results_table}
\end{minipage}
\hfill
\begin{minipage}[t]{0.37\textwidth}
\centering
{\fontsize{8pt}{10pt}\selectfont
\begin{tabular}{ r|c}
 \textbf{Brain MRIs} & DICE \\ 
  \hline
  \hline
AREPAS**  & 0.118 {\color{red}(↓61.9\%)} \\ 
AREPAS*  & 0.118 {\color{red}(↓61.9\%)} \\ 
AutoDDPM \cite{bercea2024diffusion} & 0.170 {\color{red}(↓45.2\%)} \\ 
AnoDDPM \cite{bercea2024diffusion} & 0.181 {\color{red}(↓41.6\%)} \\ 
IterMask2 & 0.216 {\color{red}(↓30.3\%)} \\ 
THOR \cite{bercea2024diffusion} & 0.297 {\color{red}(↓4.2\%)} \\ 
\hline
\textbf{AREPAS} & \textbf{0.310} {\color{blue}(↑4.4\%)} \\ 
\end{tabular}
}
\label{tab:results_table_MRI}
\end{minipage}
\end{table}

\section{Discussion}

We propose a method for anomaly detection in imaging data exhibiting fine grained anatomical detail. Results demonstrate that reconstructing real images from canny-edge maps combined with semantic patch-scoring between reconstructed and real images can increase unsupervised anomaly segmentation performance and generalizes well across imaging modalities and lesion types. 

\paragraph{Canny-edges as basis for reconstruction} In generative anomaly detection and "corrupt and reconstruct" approaches in particular, input information is restricted. In diffusion models, information is destroyed by adding noise. In IterMask one of the input channels contains a high-pass frequency filtered image, disregarding low frequency information. Edge-maps preserve fine grained structural information, but omit intensity data, fostering the network to rely on learned training distribution of intensities. Facing varying intensities, the patch similarity scoring model learns the distribution of deviations observed in normal images and is thus more robust than mere residual pixel difference calculation. Anomalies associated with structural deformations are tackled by incorporate the edge-image augmentation regime.

\paragraph{Limitations} This work does not include a detailed hyperparameter study of the augmentations applied to edges, which may further boost performance. Our framework operates on 2D slices, a 3D extension may be more suitable for volumetric data and represents a promising direction given the model's simplicity, stability, and strong performance with limited data.

\section{Conclusion}

We propose AREPAS, a novel generative two-step approach for anomaly segmentation. Results demonstrate that this method is able to detect pixel-level anomalies in data exhibiting fine-grained anatomical structure. Results consistently show the benefit of the proposed method across imaging modalities, and types of lesions. The generative training scheme using augmented data overcomes overfitting to individual canny-edges, yielding accurate reconstructions of non-anomalous regions and robust projection of anomalous areas into the non-anomalous domain. In a second step, we demonstrate that learning normal deviations between reconstructed and real images via patch-level similarity scoring enables semantic anomaly localization, and outperforms relying only on residual pixel differences. This constitutes a novel paradigm to overcome the current limitations of reconstruction-based AD methods. 
\\
\\
\textbf{Acknowledgments.} This work was funded by the Austrian Science Fund (FWF, P 35189-B - ONSET), the European Commission under the European Union's Horizon Europe research and innovation program (No. 101080302 - AI-POD, No. 101100633 - EUCAIM). It is part of AIX-COVNET and is supported by the IAEA under the ZODIAC Zoonotic Disease Integrated Action project. The financial support by the Austrian Federal Ministry of Labour and Economy, the National Foundation for Research, Technology and Development and the Christian Doppler Research Association is gratefully acknowledged.
\\
\textbf{Disclosure of Interests.} G.L. is shareholder and co-founder of Contextflow GmbH. H.P. and G.L. have received research funding by Siemens Healthineers. Other authors have no conflicts to report.

\bibliographystyle{splncs04}

\end{document}

\end{document}